# This Intestine Does Not Exist: Multiscale Residual Variational Autoencoder for Realistic Wireless Capsule Endoscopy Image Generation

Dimitrios E. Diamantis, Panagiota Gatoula, Anastasios Koulaouzidis, and Dimitris K. Iakovidis, *Senior Member IEEE*

***Abstract*— Medical image synthesis has emerged as a promising solution to address the limited availability of annotated medical data needed for training machine learning algorithms in the context of image-based Clinical Decision Support (CDS) systems. To this end, Generative Adversarial Networks (GANs) have been mainly applied to support the algorithm training process by generating synthetic images for data augmentation. However, in the field of Wireless Capsule Endoscopy (WCE), the limited content diversity and size of existing publicly available annotated datasets, adversely affect both the training stability and synthesis performance of GANs. Aiming to a viable solution for WCE image synthesis, a novel Variational Autoencoder architecture is proposed, namely 'This Intestine Does not Exist' (TIDE). The proposed architecture comprises multiscale feature extraction convolutional blocks and residual connections, which enable the generation of high-quality and diverse datasets even with a limited number of training images. Contrary to the current approaches, which are oriented towards the augmentation of the available datasets, this study demonstrates that using TIDE, real WCE datasets can be fully substituted by artificially generated ones, without compromising classification performance. Furthermore, qualitative and user evaluation studies by experienced WCE specialists, validate from a medical viewpoint that both the normal and abnormal WCE images synthesized by TIDE are sufficiently realistic.**

We acknowledge support of this work by the project "Smart Tourist" (MIS 5047243) which is implemented under the Action "Reinforcement of the Research and Innovation Infrastructure", funded by the Operational Programme "Competitiveness, Entrepreneurship and Innovation" (NSRF 2014-2020) and co-financed by Greece and the European Union (European Regional Development Fund).

This work involved human subjects or animals in its research. Approval of all ethical and experimental procedures and protocols was granted by the Ethics Committee of The University of Thessaly.

D.E. Diamantis, P. Gatoula, and D.K. Iakovidis, are with the Dept. of Computer Science and Biomedical Informatics, University of Thessaly, 35131 Lamia, Greece (email: {didiamantis, pgatoula, diakovidis}@uth.gr)

A. Koulaouzidis is with the Dept. of Surgery, Odense University Hospital and Svendborg Hospital, Denmark, and the Dept. of Clinical Research, University of Southern Denmark, Denmark (anastasios.koulaouzidis@rsyd.dk)

## I. INTRODUCTION

Gastrointestinal (GI) tract diseases constitute a significant cause of mortality and morbidity, resulting in adverse economic effects on healthcare systems [1]. Early-stage detection and precise diagnosis of pathological conditions, such as inflammations, vascular conditions, or polypoid lesions, are critical for preventing such diseases. Among the methods facilitating the screening of the GI tract, Wireless Capsule Endoscopy (WCE) [2] is one of the eminent options mainly due to its non-invasive nature. Contrary to conventional techniques, such as Flexible Endoscopy (FE), WCE is performed using a swallowable, pill-sized capsule equipped with a miniature camera. The capsule traverses throughout the GI tract recording an RGB video, which is subsequently reviewed by specialized endoscopists. Since such a video is typically comprised with more than 60,000 frames, its assessment is demanding. The evaluation of the recorded WCE videos usually requires 45-90 minutes, and it is prone to human errors even by experienced clinicians [2]. Aiming to mitigate the risk of such errors, various image-based Clinical Decision Support (CDS) systems have been proposed [3], [4] A significant aspect concerning the performance of CDS systems is their generalization ability. The availability and diversity of training data, directly impact the generalization capability of such systems. Although there are several publicly available annotated datasets for non-medical applications, many of which are quite large, *e.g.*, ImageNet [5], in the medical imaging domain, privacy regulations, such as the General Data Protection Regulation (GDPR) [6], make medical data acquisition challenging, even when their use is destined for research. Moreover, the amount of time and the cost of medical data annotation adversely contribute to their availability. Regarding WCE, the existing open annotated datasets are still limited, they are generally smaller than other datasets, often characterized by low diversity, as they contain many similar images with a narrow range of abnormality types, and in most cases, they are highly imbalanced [7], [8]. Consequently, the use of such datasets for training of contemporary deep learning-based CDS systems, limit their effectiveness for detection and characterization of abnormalities.

To address these issues, conventional image augmentation techniques, enriching the datasets with rotated, translated, and scaled versions of the training images, have been employed to enhance the generalization ability of CDS systems. More recently, approaches relying on deep neural networks have been investigated to increase the number of training images further by generating synthetic images. Generative Adversarial Networks (GANs) [9] are considered as a standard option for image synthesis tasks. In applications such as natural images or portaitsynthesis, where data availability is not an issue, the performance of GANs is remarkable [10–12]. However, in many cases, where data availability is limited, the applicability of GANs implies various training problems, including mode collapse, unconvergence and instability [13].

Current studies [14–17], indicate that still the problem of endoscopic image synthesis is far from being resolved. In most cases, synthetic images include artifacts attributed to imbalanced or insufficient training data, whereas the plausibility of the resulting synthetic images remains the main challenge.

Aiming to cope with these issues, in this study a novel approach to the generation of synthetic WCE images is proposed, based on the concept of Variational Autoencoders (VAEs) [18]. The contributions of this paper can be summarized as follows:

- It proposes a novel VAE architecture named 'This Intestine Does not Exist' (TIDE), which combines multiscale feature extraction and residual learning, to capture feature-rich representations of the input volume, enabling training on a small number of samples.
- It applies TIDE in the context of WCE image synthesis aiming to fully substitute real training sets with synthetic ones. Studies from other researchers have investigated WCE image synthesis only in the context of data augmentation, where just a subset of the training set was composed of synthetic images.
- It performs a spherical experimental evaluation study that covers quantitative and qualitative aspects, including a user evaluation study performed by WCE specialists, which verified that it is very hard to distinguish the synthetic datasets from the real ones.

Alongside this study, a demonstration website[1], has been created aiming to present the performance of the TIDE openly, and to become the first publicly available real-time intestine dataset generation platform. It is worth noting that the platform was developed using the Algorithm-agnostic architecture for Scalable Machine Learning (ASML), that we proposed in [19].

The rest of this paper is organized into four sections. Section II outlines the contribution of generative models in the medical imaging domain emphasizing the synthesis of endoscopic images. Section III presents the proposed VAE architecture for image generation. Section IV describes the evaluation methodology and includes comparative results obtained from the conducted experiments. Insights of this study are discussed in Section V, and in the last section, conclusions are drawn, and future directions are suggested.

## II. Related Work

In medical imaging, synthetic image generation has stimulated great scientific interest and several studies have been conducted in a variety of contexts, mainly using GANs. Most of the renowned GAN architectures have been applied for medical image synthesis. For instance, DCGAN [20] has been used to generate plausible brain MRI images [21]. In the spirit of conditional image generation, Mahapatra *et al*. [22] expanded the Pix2Pix framework [12] for the production of realistic-looking chest X-ray images with nodules, based on manually segmented regions. Jin *et al.* [23] expanded that framework for synthesizing 3D nodules in CT images. Another widely used adversarial framework, called PGGAN [24], was trained for the generation of convincing dermoscopic images with skin lesions [25]. The adversarial learning scheme proposed in [26], called CycleGAN, has been used for cross-modality medical image synthesis in [27], where it was applied for unpaired image-to-image translation between MRI and CT modalities of brain images. Cai *et al.* [28] modified the standard CycleGAN framework for supporting simultaneous 3D synthesis and segmentation between MRI and CT modalities of cardiac and pancreatic images, while preserving the anatomical structures. In [29], the Unsupervised Image-to-image Translation (UNIT) VAE-GAN framework [30] was applied to generate eye fundus images from segmented vessel trees. Hirte *et al*. [31] applied another variation of a hybrid VAE-GAN model [32], to generate realistic MR brain images with improved diversity.

In the field of endoscopy, generating realistic images has proved to be a more challenging task [33], [34]. This can be explained by the fact that no specific patterns are inherent in endoscopic images, nor can their content be described by well-defined structures, as in the case of CT, MRI or other medical imaging modalities. The work of [17] presented a GAN conditioned on edge-filtering combined with a mask input to synthesize images. The study focused on polyp image generation, aiming to improve polyp detection in colonoscopy videos. However, limitations of that method include non-deterministic polyp generation, and insufficient variation of generated polyp features in terms of color and texture. Various adversarial frameworks have been proposed to address the issue of imbalanced datasets by generating images of underrepresented classes in the context of endoscopic image classification. In [15], a patch-based methodology was adopted to incorporate gastric cancer findings in normal gastroscopy images. However, as commented in [14], the positioning of the polyps in that methodology is performed manually otherwise the result can be unnatural, especially with respect to the polyp features, such as color and texture. He *et al*. [14] introduced a data augmentation technique based on the GAN model of [35] by following an adversarial attack process. Lately, in the work presented in [36], the generative framework proposed in [37] was adapted to produce random polyp masks, which were then combined with normal colonoscopy images to construct a conditional input. The formulated conditional input, was leveraged for training a CycleGAN model [26] to synthesize polyp images. In [38], a dual GAN framework conditioned on polyp masks was presented for augmenting polyp findings in colonoscopy images. However, the synthesis results in both [36] and [38] depended on the positioning of the polyp masks, which were only sometimes naturally blended with healthy endoscopic images. Recently, StyleGANv2, which is a GAN architecture originally introduced for face synthesis [11], was used to enhance the training datasets for the detection of polyp lesions in endoscopic videos [34]. Although, that work produced realistic images in the context of polyp image synthesis, the reproducibility of its results is difficult as it relied on a private database with thousands of images available for training.

[1] https://this-intestine-does-not-exist.com

Most of the published studies on endoscopic image synthesis have focused on generating colonoscopic or gastroscopic images, either normal, or abnormal with polypoid lesions, whereas fewer works have investigated WCE image generation. In WCE, the images are of lower resolution, and the number of abnormal images is usually smaller, since the endoscopist cannot control the capsule endoscope to capture several frames of the lesions found, as in the case of FE. Also, WCE is more commonly applied for the examination of the small bowel, which is very difficult to be approached by FEs, and it is invaluable for the evaluation of Inflammatory Bowel Disease (IBD), and especially of the Crohn's disease (CD) [39]. Moreover, the incidence of small bowel malignancy/neoplasia – although increasing over the last decades – remains markedly lower than colorectal or gastric neoplasia [40].

In the context of WCE image generation, Ahn *et al.* [41] adapted the hybrid VAE-GAN framework originally proposed in [42], to augment an existing WCE dataset, so as to improve the generalization performance of an image-based CDS system for small bowel abnormality detection. Nevertheless, the results were only indicative, not specifying the target pathological conditions, and the synthetic images suffered from blurriness, making them easily distinguishable from the real ones.

In the more complex framework of synthesizing WCE images of the small bowel, containing various inflammatory conditions, a non-stationary Texture Synthesis GAN (TS-GAN) was presented in [43]. However, the generated images had artifacts, which degraded the quality of image synthesis. These weaknesses can be partially attributed to the limited number of training samples, since the performance of GAN models typically relies on both the quantity and the diversity of the training data. To deal with this drawback, a conventional VAE, named EndoVAE, was proposed in [44] for WCE image generation. That model was composed of convolutional layers with single-scale filters in a sequential arrangement.

In this paper we propose a different VAE architecture for WCE image generation that unlike the previous ones, it incorporates a multiscale feature extraction scheme and residual connections, aiming to provide images of enhanced quality. To the best of our knowledge such a VAE architecture has not been previously proposed.

## III. Methodology

The novel VAE architecture proposed in this study, named TIDE, uses of a series of multiscale blocks (MSBs) that extract features under multiple scales, aiming to capture a feature-rich representation of the input. It is complemented by residual connections to further enhance the feature extraction scheme, and tackle the problem of vanishing gradient, encountered in deep learning models. An MSB is illustrated in Fig. 1. The purpose of these modules is twofold. Firstly, they aim to abstract image information from an input volume at various scales. Secondly, they realize a learnable fusion of the extracted information to capture more diverse representations of the input volume. Each module receives a volume of feature maps as input, which is forwarded to three parallel convolutional layers. Each of these layers has the same number of filters, and extracts features from different scales by performing convolution operations using 3×3, 5×5, and 7×7 kernel sizes. Thus, small, medium, and large features can be extracted from the input volume. The parallel output of these layers is concatenated depth-wise, and passed forward to two additional consecutive convolutional layers, forming, in this way, a feature-rich representation of the input volume. The first convolutional layer consists of filters equal to the sum of the filters of the concatenated feature maps and it performs a pointwise convolution operation, effectively facilitating aggregation of the concatenated output. The next convolutional layer consists of the same number of filters included in the parallel convolutional layers. The activation functions of all layers mentioned are the Rectified Linear Units (ReLUs).

The entire architecture of the TIDE model is illustrated in Fig. 2. It comprises two parts: an encoding network and a decoding network. The encoder receives an input volume of RGB endoscopic images, either normal or abnormal, denoted as $x$. The proposed encoder consists of convolutional layers destined to perform pooling operations, and to extract multiscale features. Specifically, the architecture of the encoder sequentially includes a convolutional layer with 16 filters followed by four MSBs. The first MSB consists of 32 filters, which is doubled for every next module. Convolutional layers are interposed between the consecutive MSBs to perform pooling operations by reducing the size of the intermediate feature volume to half. Those layers are composed of 64 128, and 256, filters and perform convolutional operations with a kernel size of 3×3.

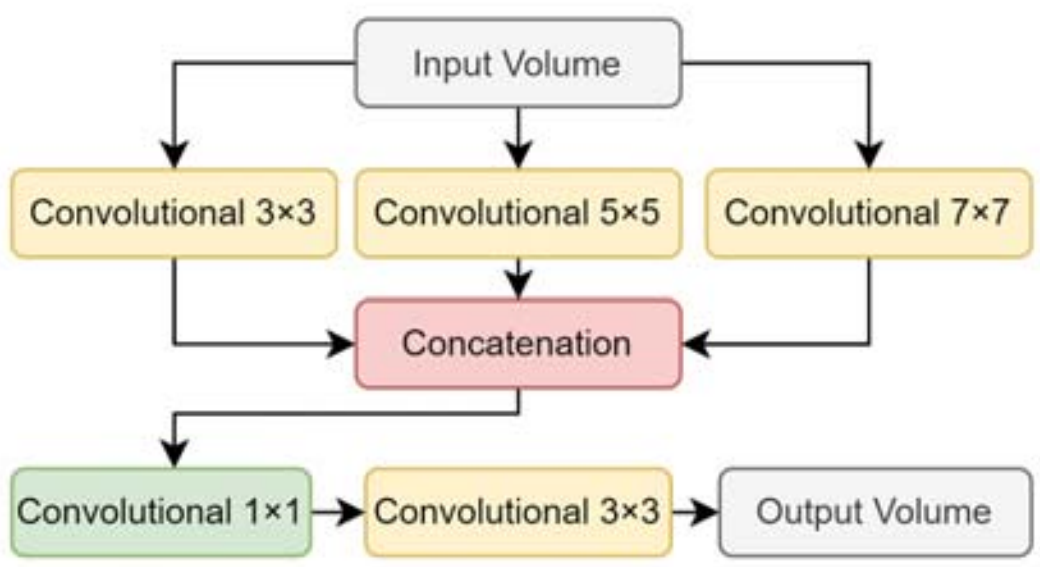

Fig. 1. Multiscale feature extraction module.

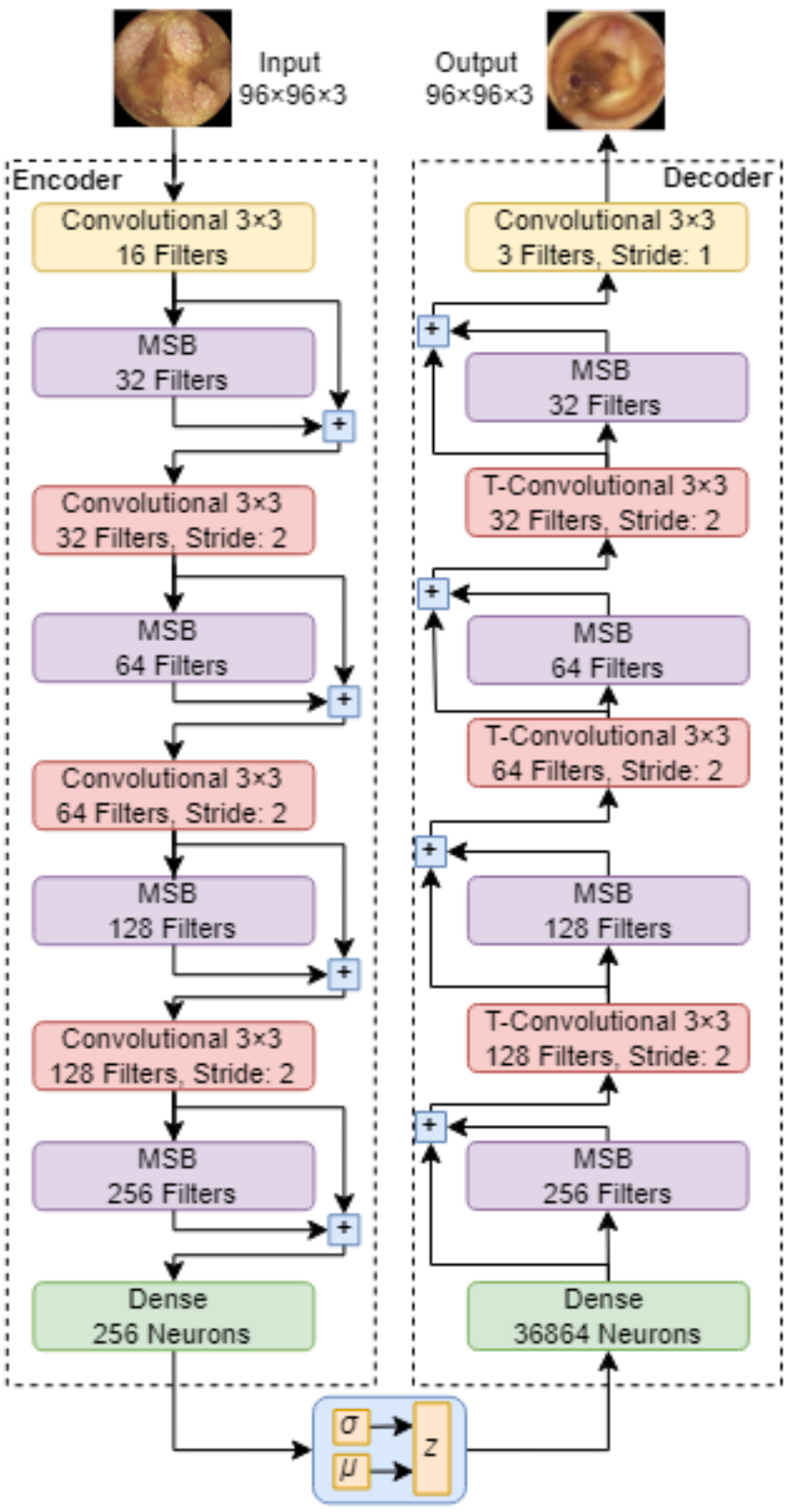

Fig. 2. TIDE architecture.

Residual connections are employed to preserve the features extracted from shallower feature extraction modules. Therefore, the input feature map volume of each module undergoes a convolutional operation with a 3×3 kernel size and a number of kernels in accordance with the number of filters leveraged by the convolutional layers of this module. The resulting feature map is aggregated with the output volume of each feature extraction module by an addition operator. Finally, the output of the residual connection is promoted to a pointwise convolution layer, preserving the number of filters of the previous convolutional layer. All the convolutional operations performed in the encoder use ReLU as an activation function.

The encoder network is tasked to compress the input volume, *i.e.*, the endoscopic images, to two different latent vectors corresponding to the statistical parameters, mean μ and standard deviation σ, of a Gaussian distribution. Therefore, the output volume of the convolutional part of the encoder is flattened and directly enters a fully connected layer with 256 neurons, followed by two separated fully connected layers connected to the previous one. Each of these two layers comprises 6 neurons with no activations that estimate the parameters $\mu$, $\sigma$ of the latent space distributions.

Following this, the decoder network randomly samples a six-dimensional vector $z$ from the distribution approximated by the encoder. Thus, the decoder, considering the latent representation $z$, reconstructs the input volume. At the top of the decoder's architecture a fully connected layer resides, having 36,864 neurons. Next, the decoder adopts the architecture of the encoder, yet with an opposite order of the MSBs that, in the case of the decoder, are separated with transposed convolutional layers for performing up-sampling of the intermediate feature volume. At the end of the decoder, a transposed convolutional layer is placed, with 3 filters to predict the reconstructed RGB input volume. The spatial dimensions of the output of the decoder correspond to those of the initial volume $x$ of endoscopic images. Consequently, the proposed VAE architecture synthesizes images of the same resolution received in the input. All the transpose convolutional operations are conducted with kernels of size 3×3 and ReLU functions as neural activations, except from the prediction layer that adopts the log-sigmoid activation function.

According to [18] the total loss backpropagated to train a VAE model can be formulated as follows:

$$\mathcal{L}(\vartheta,\varphi;\,\boldsymbol{x}_i) = \mathbb{E}_{q(\boldsymbol{z}\,|\boldsymbol{x}_i;\,\varphi)} \log p(\boldsymbol{x}_i|\boldsymbol{z}\,;\vartheta) - KL\big(q(\boldsymbol{z}\,|\boldsymbol{x}_i;\varphi)\,||\,p(\boldsymbol{z};\,\vartheta)\big) \tag{1}$$

where the first term corresponds to the reconstruction error of the decoder, and the second term approximates the Kullback-Leibler (KL) divergence. The KL-divergence is employed to ensure that the encoder compresses the input volume into a latent representation that follows a prior distribution $p(\boldsymbol{z};\vartheta)$. The prior distribution $p(\boldsymbol{z};\vartheta)$ is formulated as a multivariate Gaussian distribution $\mathcal{N}(\boldsymbol{z};0,\mathbf{I})$. We let the true intractable posterior distribution $p(\boldsymbol{z}|\boldsymbol{x}_i;\vartheta)$ be an approximation of the Gaussian with an approximately diagonal covariance that is estimated according to Eq. (2):

$$\log q(\boldsymbol{z}\,|\boldsymbol{x}_i\,;\varphi) = \log \mathrm{N}(\boldsymbol{z}\,;\boldsymbol{\mu}_i,\boldsymbol{\sigma}_i^2\,\mathbf{I}) \tag{2}$$

where $\boldsymbol{\mu}_i$ and $\boldsymbol{\sigma}_i^2$ are the outputs of the encoder part of VAE. Thus, Eq. (1) is formulated as follows:

$$\mathcal{L}(\vartheta,\varphi;\,\boldsymbol{x}_i) \simeq \frac{1}{2}\sum_{j=1}^{J}\left(1+\log\sigma_{i,j}^2-\sigma_{i,j}^2-\mu_{i,j}^2\right) + \frac{1}{L}\sum_{L=1}^{L}\log \mathrm{p}(\boldsymbol{x}_i|\boldsymbol{z}_l\,;\vartheta) \tag{3}$$

where

$$\boldsymbol{z}_l = \boldsymbol{\mu}_i + \boldsymbol{\sigma}_i \odot \boldsymbol{\epsilon}_l \text{ and } \boldsymbol{\epsilon}_l \sim \mathcal{N}(0,\mathbf{I}) \tag{4}$$

$J$ denotes the dimensionality of the underlying manifold, $L$ refers to the sample size of the Monte Carlo method sampling from the approximate posterior distribution of the encoder, $\varphi$ represents the parameters of the encoder network, $\vartheta$ represents the parameters of the decoder network, and, and symbol $\odot$ represents the Hadamard product operation.

## IV. Experiments and Results

### A. Datasets and Training of the Generative Model

Considering the clinical utility of WCE and its importance for evaluating inflammatory conditions of the small bowel, two datasets, namely KID Dataset 2, and the Kvasir-Capsule dataset, were used for experimentation [45], [46]. To the best of our knowledge these datasets are the only publicly available annotated WCE datasets that include inflammatory lesions, such as erythemas, erosions, and ulcers. KID [45] is a WCE database designed to assess CDS systems. It includes 728 normal images and 227 images with inflammatory lesions of the small bowel, with a resolution of 360×360 pixels. The images were acquired using Mirocam® (IntroMedic Co., Seoul, Korea) capsule endoscopes. The Kvasir-Capsule dataset [46] includes a total of 34,338 normal images and 1,519 images of inflammatory lesions, with a resolution of 336×336 pixels. The images were acquired using an Olympus EC-S10 endocapsule. Despite the relatively larger size of this dataset, it contains many images that are similar to each other. To reduce frame redundancy, while maintaining the two datasets equivalent in size, a subset of 728 normal and 227 abnormal representative, non-overlapping WCE images from the whole dataset, was sampled using the image mining methodology described in [47] (the filenames of the sampled images are provided in the supplementary material of this paper). Different TIDE models were trained separately, on normal, and abnormal subsets of the KID and Kvasir-Capsule datasets, respectively, *i.e.*, a TIDE model was trained to generate normal images, and another one was trained to generate abnormal images, per dataset. The two datasets were not considered jointly because they have been acquired using different types of capsule endoscopes. Regarding the training process, no other data augmentation techniques were applied on the training sets. TIDE was trained using early stopping, with a maximum limit of 5,000 epochs, using batches of 128 samples. Considering its effectiveness in relevant applications, the Adam optimizer [48] was selected to train the model, using a learning rate initially set to 0.001.

### B. Quantitative Evaluation

The main goal of this experimental study is to investigate if a WCE dataset composed solely of synthetic images can be effectively used to train a classifier, so that it accurately learns to discriminate real abnormal from real normal images. Therefore, the classification performance can be considered as a representative index for quantitative evaluation of the synthetic WCE datasets generated using TIDE [49]. The classification performance was quantified by examining the Receiver Operating Characteristic (ROC) curves, considering that the WCE datasets are imbalanced, and that the classification problem is binary. ROC curves indicate the diagnostic ability of a classification system by illustrating a tradeoff between True Positive (TPR) and False Positive (FPR) Rates using various decision thresholds. The Area Under ROC (AUC) measure [50], was computed to assess the performance of the trained classification models, because, unlike other measures, such as accuracy, sensitivity and specificity, which are obtained using only a single decision threshold, it is insensitive to imbalanced class distributions [50], [51].

Considering that the extensive experimental work required for this study is computationally demanding, we selected LB-FCN *light* classifier [52], as a computationally more efficient version of LB-FCN, which is a state-of-the-art classifier proposed for accurate classification of endoscopic images [53].

The experimental procedure can be outlined as follows: 1) a reference performance of LB-FCN *light* per real dataset was estimated for the classification of each dataset into normal and abnormal (inflammation) classes; 2) the trained TIDE models were used to randomly generate different sets of synthetic normal and abnormal WCE images; 3) for each dataset, the same LB-FCN *light* classifier was trained solely on the synthetic normal and abnormal images, and tested, exclusively on the respective real images.

TABLE I
CLASSIFICATION RESULTS (AUC %) ON REAL AND SYNTHETIC IMAGES

| | *KID* [45] | *Kvasir-Capsule* [46] |
|---|---|---|
| *Real* | 90.9 ± 0.8 | 80.0 ± 0.7 |
| *TS-GAN* [43] | 79.1 ± 0.7 | 68.8 ± 0.7 |
| *CycleGAN* [26] | 62.9 ± 1.2 | 72.1 ± 1.1 |
| *StyleGANv2* [11] | 61.7 ± 1.1 | 71.8 ± 1.0 |
| *EndoVAE* [44] | 81.9 ± 0.9 | 71.3 ± 0.3 |
| *TIDE* | **89.4 ± 1.2** | **80.2 ± 0.6** |

Aiming to a fair comparison between the classification performance results obtained using the KID and Kvasir-Capsule datasets, the same number and proportion of normal and abnormal images was considered (*i.e.*, 728 normal and 227 abnormal synthetic images). In all experiments, a stratified 10-fold cross validation approach was adopted to alleviate a potential selection bias.

More specifically, both the synthetic dataset and the dataset with the real images were split into ten subsets that were fully disjoint, from which, nine subsets of the synthetic dataset were used for training, and a subset from the real dataset was used for testing. The process was repeated ten times, selecting different training and testing subsets, until all the real subsets were used for testing. The training settings of LB-FCN *light* are the ones suggested in [44]. Considering that the generation of the synthetic images was random, the whole experimental procedure was also repeated ten times and average results with standard deviations were recorded.

The results of the quantitative experimental evaluation of the TIDE model, were compared with the results obtained by relevant state-of-the-art models for WCE image generation, *i.e.*, TS-GAN [43], and EndoVAE [44]. The respective classification performances are summarized in Table I, which also includes the reference results obtained per dataset using the real images. It can be noticed that TIDE offers an improved classification performance over all the compared models, and more importantly, it is comparable to the reference one. This validates the hypothesis that real training images can be substituted by synthetic ones, without sacrificing classification performance.

It should be noted that we have also experimented using methods that have been previously applied for the generation of FE images, including CycleGAN [26] and StyleGANv2 [11]. Training these architectures was challenging, mainly because of problems deriving from the small number of the available WCE training samples. Such issues include, low-quality image generation, lack of diversity, presence of artifacts and more importantly mode-collapse [54]. Initially we tried to take advantage of pre-trained models (on ImageNet [5]) for weight initialization. However, both networks could not converge, resulting in mode collapse, early on training. Training CycleGAN and StyleGANv2 from scratch, resulted in images that, while in some cases resembled WCE images, they were mostly unnatural, with noise artifacts being prevalent. This is reflected in the lower classification performance obtained using the datasets generated by CycleGAN and StyleGANv2, as indicated in Table I.

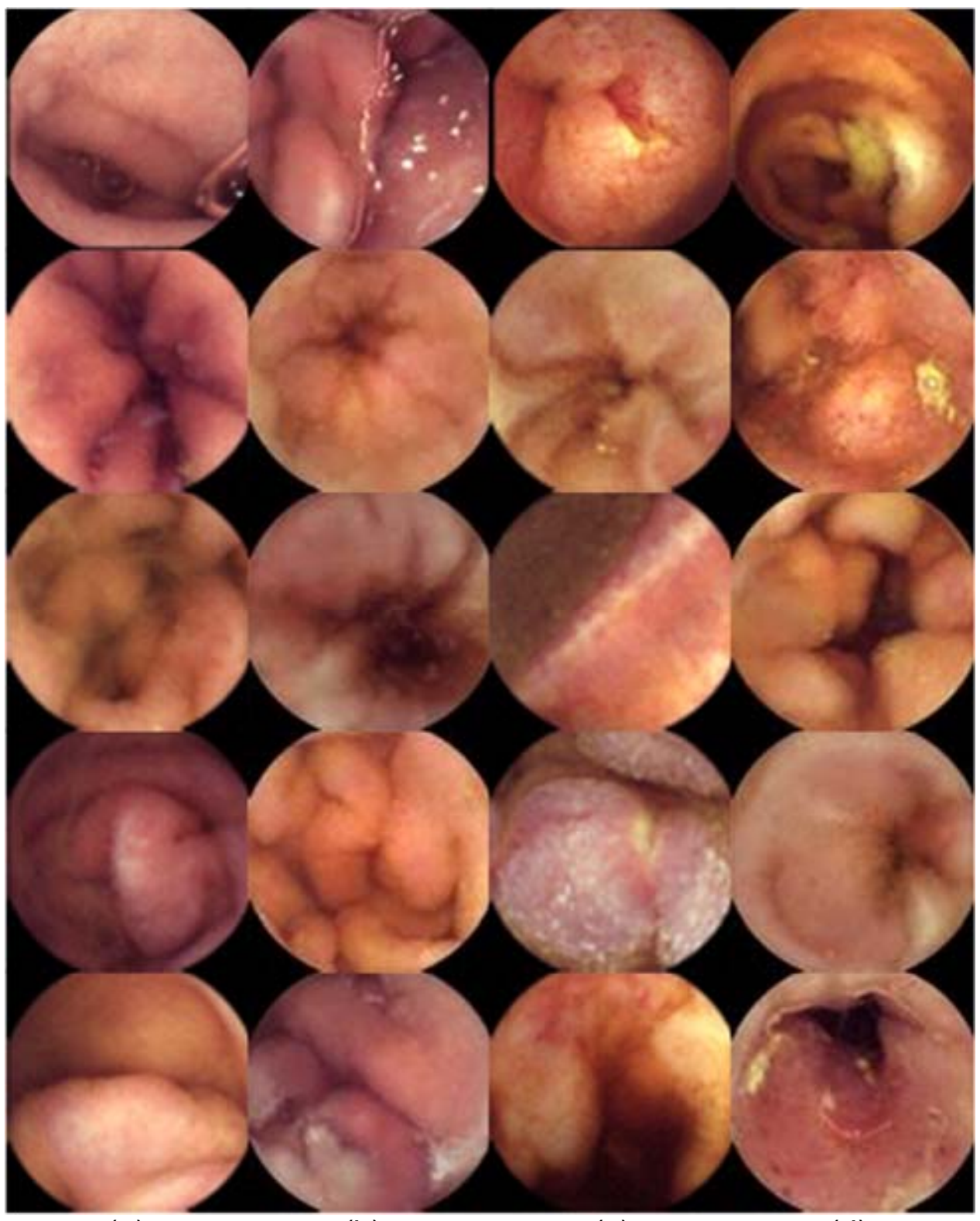

(a) (b) (c) (d)

Fig. 4. Real and synthetic endoscopy images illustrating small bowel tissue from KID dataset. (a) Real normal images. (b) Normal images generated by TIDE.(c) Real abnormal images. (d) Abnormal images generated by TIDE.

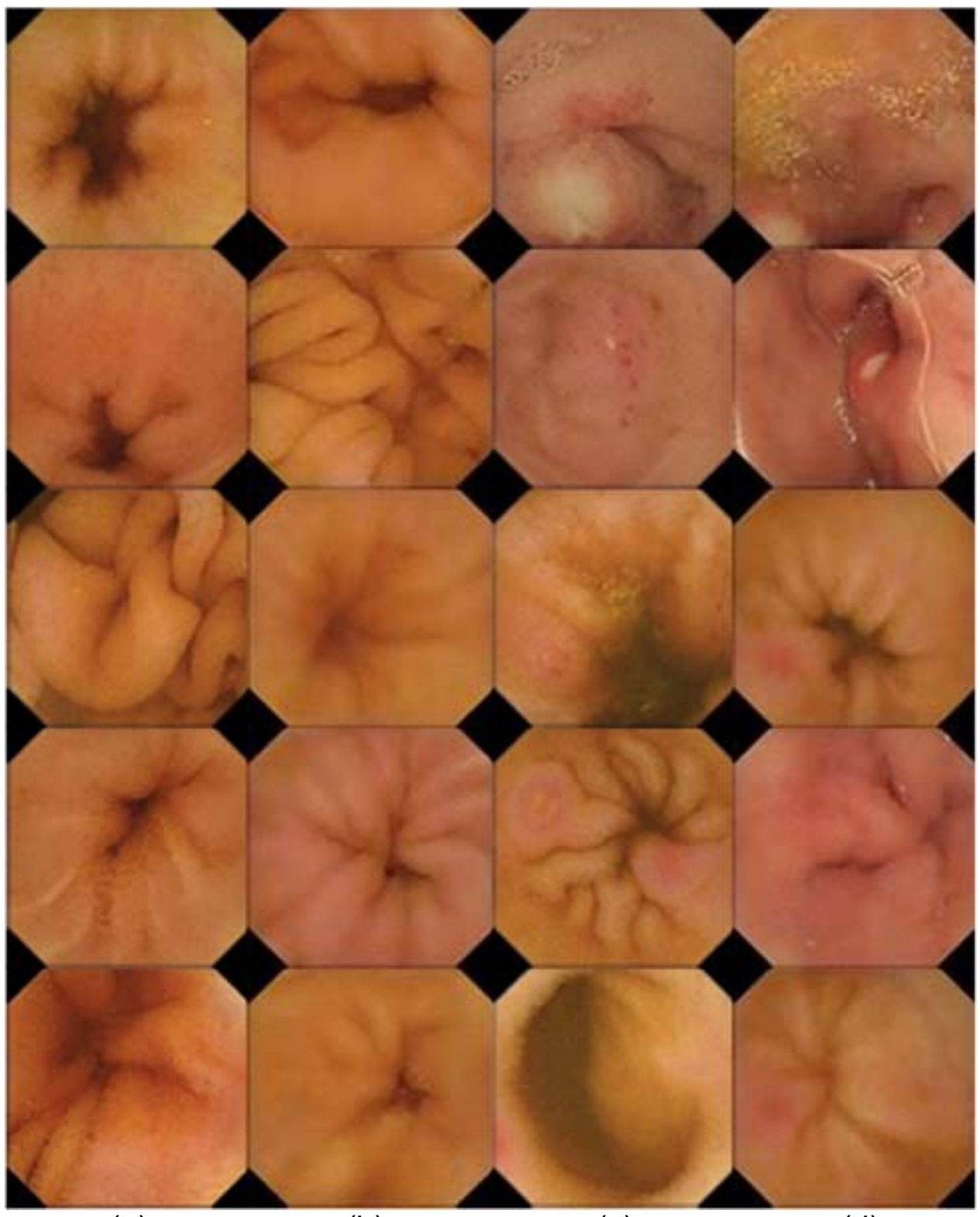

(a) (b) (c) (d)

Fig. 5. Real and synthetic endoscopy images illustrating small bowel tissue from Capsule-Kvasir dataset. (a) Real normal images. (b) Normal images generated by TIDE.(c) Real abnormal images. (d) Abnormal images generated by TIDE.

### C. Qualitative Comparison

A qualitative, visual comparison, between the real images of the KID and Kvasir-Capsule datasets, and representative images generated by TIDE, can be performed by examining the images of Fig. 4 and Fig. 5, respectively. These figures show that TIDE generates realistic endoscopic images with a diversity resembling that of the real images. Particularly, it can be noticed that in the synthetic images generated by TIDE, the visible characteristics of the real tissues are preserved, including color, texture and shape. The lesions generated in the case of the abnormal images, not only look like the ones in the real abnormal images, but they are also naturally positioned and blended with the normal tissue. Furthermore, the generated images include realistically reproduced bubbles and debris, which are common in real WCE images.

Figures 6 and 7 provide a comparative visualization of normal and abnormal images, respectively, generated using different state-of-the-art generative models, namely TS-GAN [43], CycleGAN [26], StyleGANv2 [11], and EndoVAE [44]. More specifically, with respect to the synthetic normal images in Fig. 6, from a medical viewpoint, TS-GAN (Fig. 6(a)) provides rather realistic-looking synthetic images but with a marked granularity of the image and pixelation that is not common in the usual small-bowel capsule endoscopy images; CycleGAN (Fig. 6(b)), provides an entirely pixelated set of images that, together, the degree of haziness, does not allow any safe observations to be performed with that set; the images generated by StyleGANv2 (Fig. 6(c)) compared with CycleGAN, represent a much-improved version but still suffer from marked image haziness and an outcome that points toward a non-realistic set of normal small bowel images; EndoVAE (Fig. 6(d)) generated a set of realistic normal small-bowel images, which, despite the marked improvement as compared with the results of the previous models (even with TS-GAN), the amount of haziness and the presence of ultra-white artefacts give it away as a non-real dataset; the images generated by TIDE (Fig. 6(e)) are characterized by clarity and higher definition as compared to the previous ones, with only scarce presence of artefacts; the additional water/air bubble interface helps in providing extra realistic features.

Regarding the abnormal images in Fig. 7, all GAN-based methods (Fig. 7(a-d)) show non-realistic abnormalities of the small bowel. Although TS-GAN is an early model for WCE image generation, it provides images with a rather clear impression of possible mucosal infiltration/induration by a relevant process. However, the images lack clarity, including some artifacts, and they cannot be used to deduct diagnostic conclusions. On the contrary, the last column includes images of mucosal ulceration, characterized by a realistic texture that approximates that of real images and they can be used for clinical training and other functions.

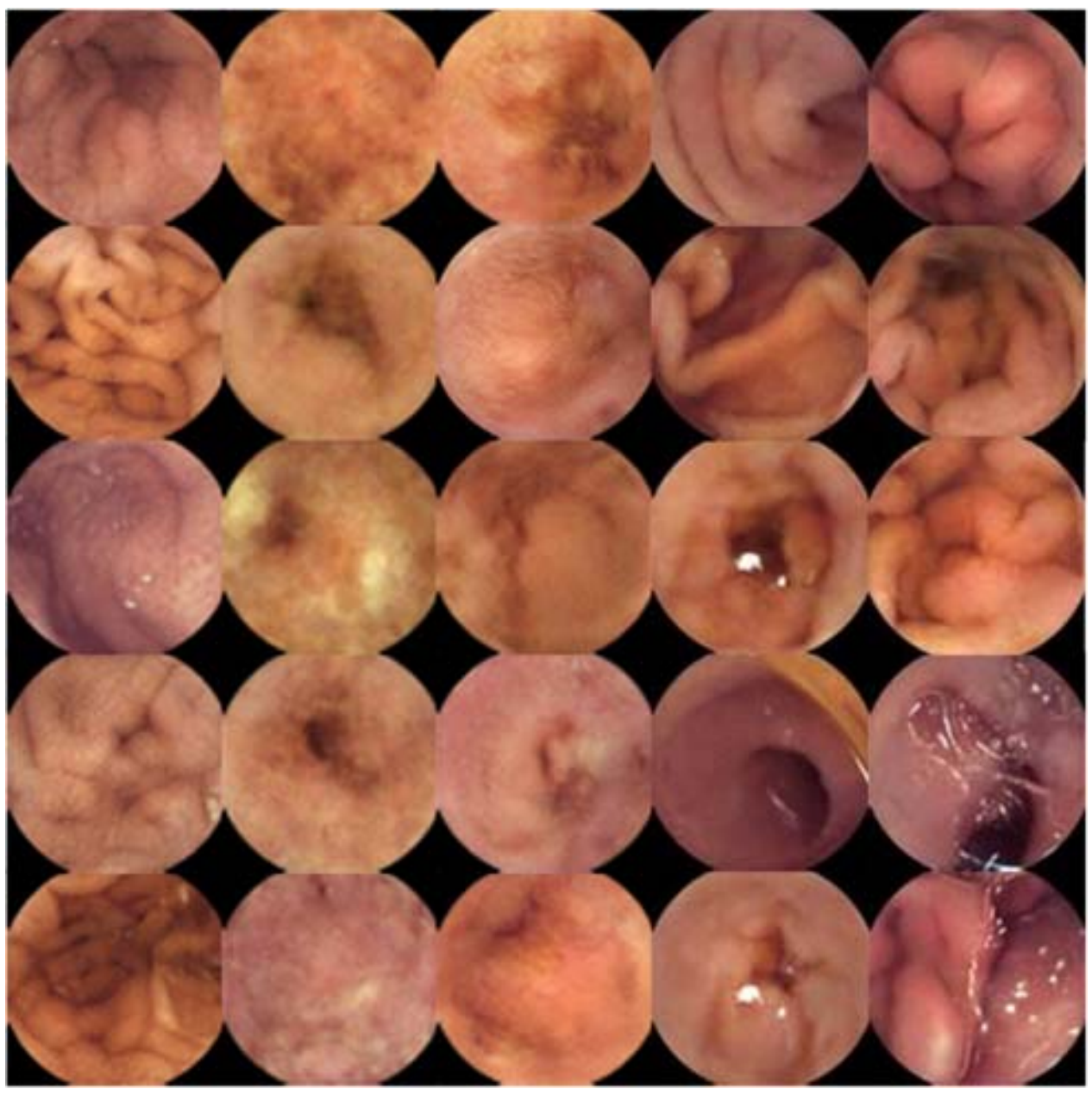

(a) (b) (c) (d) (e)

Fig. 6. Synthetic normal WCE images produced by different generative models. (a) TS-GAN [43]. (b) CycleGAN [26]. (c) StyleGANv2 [11]. (d) EndoVAE [44]. (e) TIDE.

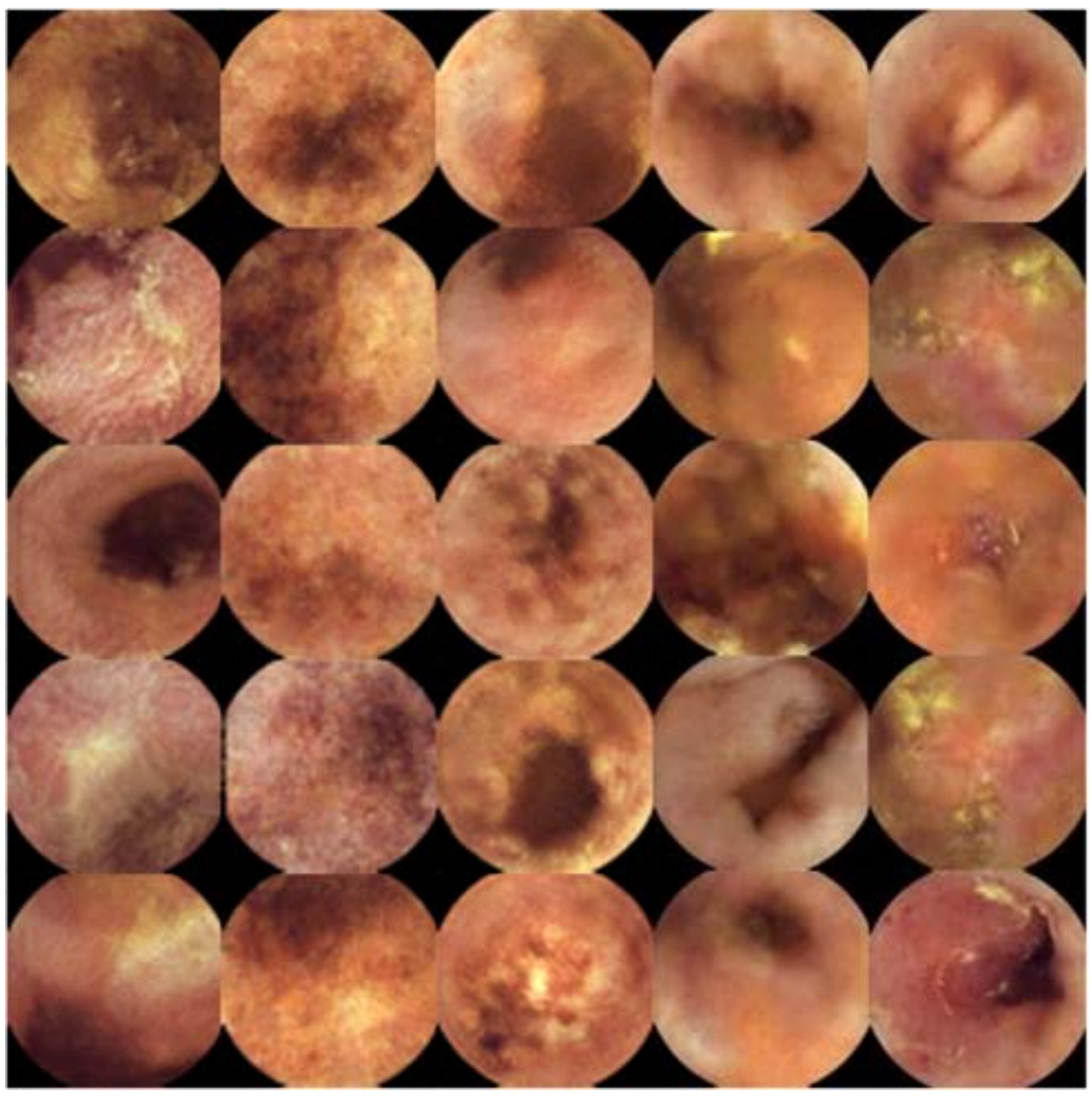

(a) (b) (c) (d) (e)

Fig. 7. Synthetic abnormal WCE images produced by different generative models. (a) TS-GAN [43]. (b) CycleGAN [26]. (c) StyleGANv2 [11]. (d) EndoVAE [44]. (e) TIDE.

To validate the visual observations with respect to the diversity of the generated images a complementary experimental study was conducted. The exponential of the Shannon entropy of the eigenvalues of a kernel similarity matrix was considered as a generic, domain-independent measure [55]. Let us consider a collection of independent samples $x_1, x_2, \ldots, x_n \in \mathcal{X}$ , $\mathbf{K} \in \mathbb{R}^{n \times n}$ a positive semi-definite kernel similarity matrix with $\mathrm{K}_{i,i} = 1 \, for \; i \in \{1, \ldots n\}$, and $\boldsymbol{\lambda} = (\lambda_1, \lambda_2, \ldots, \lambda_n)$ a vector with the eigenvalues of $\mathbf{K}/n$. The diversity of this collection of samples can be defined as:

$$\delta = exp\left( - \sum_{i=1}^{S} \lambda_i \, \log \lambda_i \right) \quad (5)$$

Different kernel similarity matrices can be utilized to capture the visual or semantic similarity of the samples to be evaluated. In this work both pixel-based and feature-based similarity kernels were considered, aiming to quantify the diversity with respect to the image details and semantic content, respectively. Pixel-based similarity is measured as the cosine similarity between pixel vectors and it captures differences related to low-level image features, such as the brightness and colour of the images compared. The feature-based similarity is calculated as the cosine distance between high-level features of the images. The last pooling layer of an Inception-v3 model trained on ImageNet was selected as a perceptually-relevant feature extractor [55], which has also been effectively applied in the context of WCE image representation [56]. Ultimately, the diversity $\delta_g$ of a generated dataset should be approximately equal with the diversity $\delta_r$ of the respective real dataset, *i.e.*, $\delta_{g=}\delta_r$. We consider the relative diversity, as a more meaningful measure defined as $\tilde{\delta} = \delta_g/\delta_r$, because it provides a diversity score that is independent from the diversity of the real dataset used to train the generative model, enabling direct comparisons among different datasets; therefore, this measure is maximized for $\delta_{g=}\delta_r$, *i.e.*, $\tilde{\delta} = 1$. Figure 8 illustrates the relative diversity over all the datasets generated in this study. Considering the feature-based similarities, in that figure it can be noticed that those generated by the two VAE-based models and TS-GAN have higher relative diversities than CycleGAN and StyleGANv2, whereas considering the pixel similarities, the results of all models are comparable to each other, except from the abnormal dataset generated ST-GAN, which exhibits a significantly higher relative diversity. However, TIDE outperforms EndoVAE with respect to the feature-based diversity observed in the normal datasets.

## D. User Evaluation

Based on the above, the datasets generated by TIDE result in a classification performance that is equivalent to that obtained using the respective real WCE images. Also, the images of the TIDE datasets have both a more realistic appearance and they are clearer than the datasets produced by the compared generative models. To investigate if the TIDE datasets are sufficiently realistic also for endoscopists specialized in WCE, a series of Visual Turing Tests (VTTs) was conducted. More specifically, three tests were performed using normal images and images depicting inflammatory conditions from the small bowel, automatically generated by TIDE. The first and second tests included 55 images each, with the first one having only real images from the KID database and the second one containing only synthetic ones generated by the TIDE architecture. The last test combined the two, including 110 images in total. The tests were then given to three endoscopists with 10 to 30 years of experience who were called to distinguish the synthetic from the real images. It is important to note that to avoid any selection bias, the proportion of the real and fake images in the all the tests was undisclosed to the participated experts. Additionally, the outcomes of each VTT were not announced to them until the completion of this study.

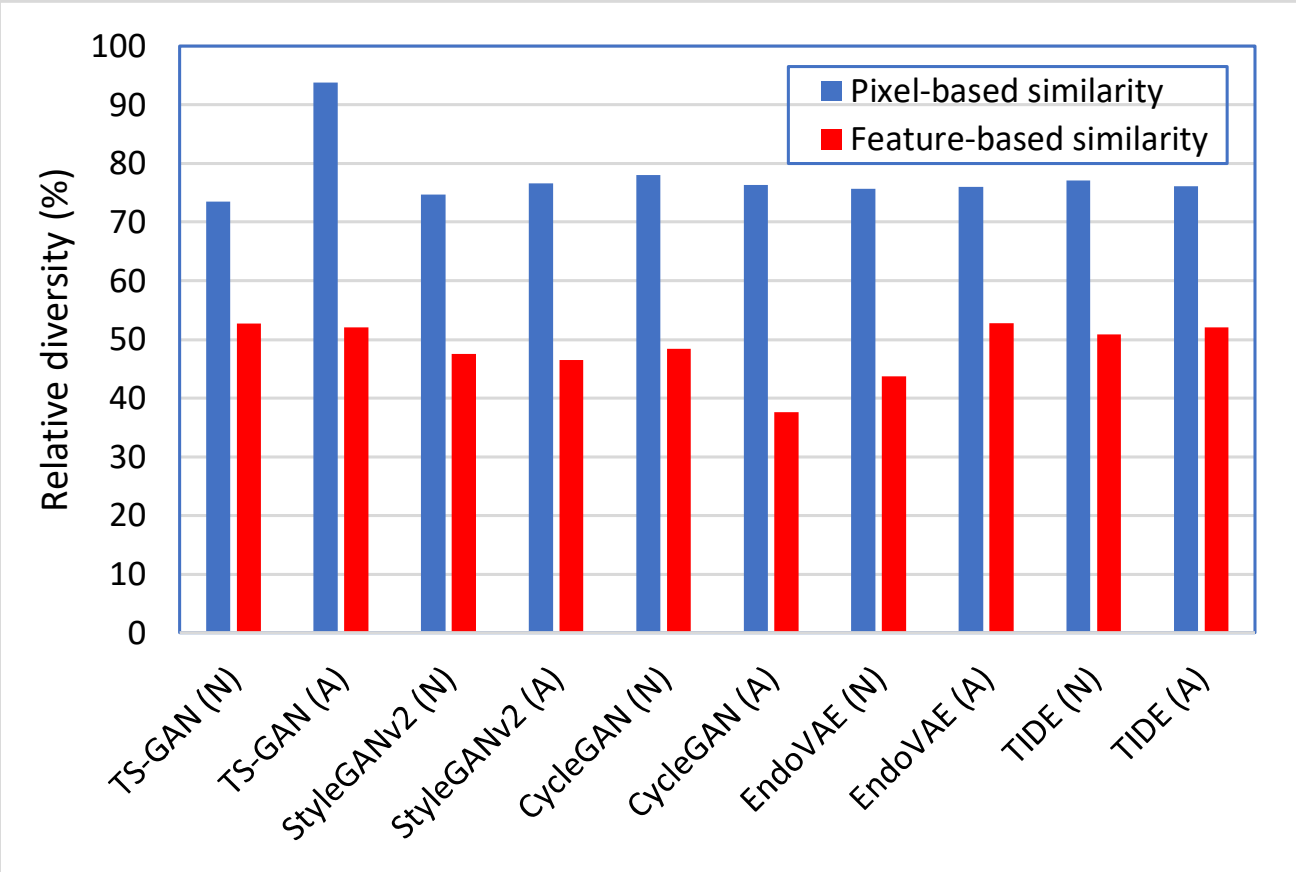

Fig. 8. Relative diversity, based on pixel and feature similarity kernels, of the normal (N) and the (A)bnormal samples of all the synthetic datasets produced by different generative models.

Table II summarizes the results of all the VTTs conducted by the three endoscopists. In the first VTT consisting of only artificially generated images produced by the TIDE architecture, the average accuracy obtained by the endoscopists was 46.1±7.3% (ranging between 38.2% and 52.7%). For the second VTT, which consists of only real images, the mean accuracy was 66.1±8.6% (ranging between 56.4% and 72.7%). The third VTT contained real and artificially synthesized images generated by the proposed methodology; the average obtained accuracy was 50.0±1.8% (ranging between 48.2% and 51.8%). Considering the real images as positive predictions and the synthetic ones as negative predictions, the mean sensitivity and specificity were 65.5±10.1% (ranging between 56.4% and 76.4%) and 34.6±6.5% (ranging between 27.3% and 40.0%), respectively (Table III). The above results validate that the endoscopic images generated by TIDE are hard to distinguish from the real ones.

TABLE II
SUMMARY OF ALL VISUAL TURING TEST RESULTS

| | Visual Turing Tests | | |
|---|---|---|---|
| | 1st VTT | 2nd VTT | 3rd VTT |
| | *Accuracy (%)* | *Accuracy (%)* | *Accuracy (%)* |
| *Endoscopist I* | 38.2 | 72.7 | 50.00 |
| *Endoscopist II* | 47.3 | 69.1 | 51.8 |
| *Endoscopist III* | 52.7 | 56.4 | 48.2 |
| *Mean* | 46.1 ± 7.3 | 66.1 ± 8.6 | 50.0 ± 1.8 |

TABLE III
MEAN SENSITIVITY AND SPECIFICITY IN THE CASE OF THE 3RD VTT

| | 3rd VTT | |
|---|---|---|
| | *Sensitivity (%)* | *Specificity (%)* |
| *Endoscopist I* | 63.6 | 36.4 |
| *Endoscopist II* | 76.4 | 27.3 |
| *Endoscopist III* | 56.4 | 40.0 |
| *Mean* | 65.5 ± 10.1 | 34.6 ± 6.5 |

## V. DISCUSSION

The limited availability of annotated datasets in medical imaging, is a barrier for essential progress in developing image-based CDS systems. Particularly in the domain of WCE the need for such progress is urgent, as the diagnostic yield remains low, and WCE specialists reach their limits by trying to maintain their concentration undistracted while examining several thousands of images [3]. This study was motivated by the need for publicly available benchmarking WCE datasets that will trigger a productive competition among image analysis researchers to effectively improve their methods for use in clinical practice. We proposed a multiscale residual VAE architecture capable of generating synthetic WCE images, and showed, using publicly available datasets, that such images can replace the real ones for training machine learning systems for the detection of abnormalities. The range of abnormalities that can be found in the small bowel, where WCE is mainly applicable, is broad. As a proof-of-concept, this study focused on inflammatory lesions, which represent a range of abnormalities associated with diseases, such as IBD and CD, affecting millions of individuals worldwide [1].

During the last decade, image generation methods have been proposed in various medical and non-medical domains. GANs and their variants have had a tremendous success mainly in generating synthetic images of human faces, and several studies have reported exceptional results in the generation of medical images [21], [23], [27], [28]. However, important factors of success in these studies constitute the large number of diverse training data, and the relatively aligned content of the training images, *e.g.*, the face images are aligned with respect to the facial features, and CT or MRI images can be aligned with respect to the depicted body structures. On the other hand, the generation of synthetic endoscopic images of the GI tract is more challenging, since their content is more diverse without features that could be considered for alignment. Related studies (Section II) have reported results based on significantly larger, usually not publicly available, training sets. These studies have also indicated issues with respect to the application of GAN-based generative models. For example, it is worth noting that [38] reports that a contemporary classification system, trained with images generated by a GAN model, reached a saturation in performance improvement after a certain point, even if more synthetic polyp images were added to the training set. This was attributed to the fact that the GAN model was unable to introduce new unseen features. The GAN was only manipulating the existing features in the training set, trying to reuse the same set of features to generate new-looking synthetic polyps. This is a common limitation of GAN-based image generation models, and it could also justify the results of TS-GAN in our study. Although that model managed to generate more plausible images than the other compared GAN models, with a relatively high diversity, the generated dataset was not sufficient to provide a classification performance equivalent to that of the respective real datasets.

The classification performance using synthetic datasets generated from the KID database, was generally higher than that observed using synthetic datasets generated from the Kvasir-Capsule dataset, regardless the type of the generation model. This could be attributed to the fact that the real Kvasir-Capsule images generally include smaller lesions than those of the real KID dataset.

The results obtained from the two VAE architectures compared in this study, namely EndoVAE and TIDE, indicate that although they can both generate quite realistic images, the improved diversity, and the higher definition of the depicted structures in the images generated by TIDE, play a significant role in the improvement of the classification performance.

In practice, TIDE could be used to generate synthetic annotated WCE datasets based on real, anonymized datasets, and the real datasets can be securely kept, within the premises of the healthcare provider. The synthetic datasets can be shared publicly with the (technical) research community, without raising any legal or ethical concerns, since they are not real, originating from statistical processing that does not allow identification of any personal information.

## VI. Conclusions

This paper presented TIDE, a novel VAE architecture for the generation of synthetic WCE images that incorporates multiscale feature extraction and residual learning in a deep learning model. A proof-of-concept case study was investigated addressing the generation of normal and abnormal images of the small bowel in the context of image-based CDS systems for the detection of inflammatory small bowel lesions. The results of the experimental evaluation of TIDE lead to the following main conclusions about the proposed architecture:

- It enables the generation of synthetic images of enhanced clarity and diversity, suitable to fully substitute real training sets for WCE image classification.
- It accomplishes effective and realistic WCE image synthesis even using a limited number of training samples.
- The synthetic images generated by TIDE are difficultly distinguished even by experienced WCE specialists.

Future research directions include the application of the proposed framework for generating images from the entire GI tract with various abnormalities and pathological conditions. Furthermore, the generality of the proposed makes it a candidate solution for generating synthetic images of other medical imaging modalities.

## Acknowledgment

We would like to thank the anonymous endoscopists who participated in our study.